\documentclass[letterpaper, 10 pt, conference]{ieeeconf}  

\IEEEoverridecommandlockouts                              
\usepackage{cite}
\usepackage{amsmath,amssymb,amsfonts}
\usepackage{algorithmic}
\usepackage{graphicx}
\usepackage{textcomp}
\usepackage{xcolor}
\usepackage{verbatim}
\usepackage{tabularx}
\usepackage{cite}
\usepackage{makecell}
\usepackage{mathtools}
\usepackage{commath}
\usepackage{array,multirow}
\usepackage[normalem]{ulem}

\newcolumntype{C}{>{\hsize=\dimexpr2\hsize+8.25\tabcolsep+\arrayrulewidth\centering\relax}X}
\newcolumntype{U}{>{\hsize=\dimexpr2\hsize+6\tabcolsep+\arrayrulewidth\centering\relax}X}
\newcolumntype{Q}{>{\hsize=\dimexpr1\hsize+1\tabcolsep+\arrayrulewidth\centering\relax}X}
\newcolumntype{S}{>{\hsize=\dimexpr1\hsize+1.1\tabcolsep+\arrayrulewidth\centering\relax}X}
\newcolumntype{O}{>{\hsize=\dimexpr1\hsize+18\tabcolsep+\arrayrulewidth\centering\relax}X}
\newcolumntype{P}{>{\hsize=\dimexpr1\hsize+9\tabcolsep+\arrayrulewidth\centering\relax}X}
\newcommand{\mc }[1] {\multicolumn{1}{Q}{\footnotesize#1}}
\newcommand{\mcs}[1] {\multicolumn{1}{C}{\small#1}}

\newcolumntype{Y}{>{\centering\arraybackslash}X}

\title{\LARGE \bf
Identity and Posture Recognition in Smart Beds \\with Deep Multitask Learning
}

\author{Vandad Davoodnia$^{1}$ and Ali Etemad$^{2}$
\thanks{$^{1}$V. Davoodnia is with the Department of Electrical and Computer Engineering, Queen's University, Kingston, ON, Canada
        {\tt\small vandad.davoodnia@queensu.ca}}%
\thanks{$^{2}$A. Etemad is with Department of Electrical and Computer Engineering, Queen's University, Kingston, ON, Canada
        {\tt\small ali.etemad@queensu.ca}}%
}

\begin{document}

\maketitle
\thispagestyle{empty}
\pagestyle{empty}

\begin{abstract}
Sleep posture analysis is widely used for clinical patient monitoring and sleep studies. Earlier research has revealed that sleep posture highly influences symptoms of diseases such as apnea and pressure ulcers. In this study, we propose a robust deep learning model capable of accurately detecting subjects and their sleeping postures using the publicly available data acquired from a commercial pressure mapping system. A combination of loss functions is used to discriminate subjects and their sleeping postures simultaneously. The experimental results show that our proposed method can identify the patients and their in-bed posture with almost no errors in a 10-fold cross-validation scheme. Furthermore, we show that our network achieves an average accuracy of up to 99\% when faced with new subjects in a leave-one-subject-out validation procedure on the three most common sleeping posture categories. We demonstrate the effects of the combined cost function over its parameter and show that learning both tasks simultaneously improves performance significantly. Finally, we evaluate our proposed pipeline by testing it over augmented images of our dataset. The proposed algorithm can ultimately be used in clinical and smart home environments as a complementary tool with other available automated patient monitoring systems. 
\end{abstract}

\section{Introduction}
Sleep is a natural activity, which is essential for life and physical well-being. It has been shown that sleeping affects symptoms of many diseases and plays an important role in systemic physiology, including metabolism and the functions of the immune, hormonal, and cardiovascular systems \cite{consensus2015joint,altevogt2006sleep}. As such, recent studies have shown in-bed postures have a high influence on occurrence of sleep diseases such as apnea \cite{lee2015changes}, pressure ulcers \cite{black2007national}, and even carpal tunnel syndrome \cite{mccabe2011preferred, mccabe2010evaluation}. For instance, sleep apnea patients are recommended to only sleep in the side-positions \cite{cartwright1984effect}. Accordingly, several studies have attempted to build in-bed pose estimation systems using data collected from pressure maps as an indicator of sleep quality \cite{seo2004intelligent, grimm2011automatic}.

Currently, aside from visual inspections and reports from patients, sleep studies mainly rely on using either physiological signals and polysomnography (PSG) \cite{blood1997comparison} or camera-based systems, which are both performed exclusively in hospital environments where patients are required to stay overnight. To address such problem, automated in-bed integrated systems are needed for clinical and smart home settings. Recently, commercial pressure sensing mattresses have been used to continuously measure pressure distribution of the body parts while the user lies on the bed \cite{pouyan2017pressure,PhysioNet}. These pressure sensing arrays can report valuable information whether the subject is lying, sitting, or standing \cite{damousis2008unobtrusive, azimi2017breathing, javaid2017balance}, which can be complementary to conventional systems. For instance, pressure maps have been used for in-bed body pose estimation \cite{liu2017bed} or measurement of physiological signals such as heart rate and respiration\cite{kortelainen2012multichannel}. There has also been increased attention on studies regarding the effects of sleeping postures on sleep apnea \cite{cartwright1984effect, penzel2001effect} and ulcer prevention using advanced monitoring technologies \cite{ostadabbas2014bed, mcinnes2015support} for applications in smart beds.

In addition to classification of postures, smart beds can enable automated recognition and identification of users, which will in turn allow for security and authentication applications, as well as personalization of smart-home experiences. For example, textile-based pressure sensors were used to perform user classification in \cite{pouyan2017pressure}.

Generally, pose detection techniques are divided into categories based on the input modality such as video and depth cameras, electrode-based sensors, or pressure mattresses. Camera-based approaches use a variety of cameras such as motion capture \cite{etemad2014classification}, range \cite{achilles2016patient}, or RGB \cite{cao2018openpose}. These methods suffer from occlusions, lightening conditions, as well as viewpoint angle and calibration effects. As a result, in-bed camera-based approaches often need to shift their attention to addressing these problems \cite{liu2017bed, wang2010robust}. Furthermore, violation of user privacy is another issue that has contributed to the lack of bed-based image datasets available for public use. Where such datasets have been available, deep learning methods have been explored for pose estimation. For instance \cite{achilles2016patient} trained a deep model to infer body pose from RGB-D data given ground truths from an optical motion capture system.

In contrast to the aforementioned approaches, pressure-based pose detection systems avoid problems such as occlusion, lightening variations, and user privacy, making them a good candidate for smart homes and clinical settings. However, Pressure mapping systems suffer from noisy measurements and the need for frequent calibration. Reference \cite{pouyan2017pressure} performed subject recognition and posture classification in three standard postures, namely, right, supine, and left. They extracted eighteen statistical features from the pressure distribution and fed them to a dense network, pre-trained by incorporating a restricted Boltzmann machine. Similarly, others have taken multimodal approaches \cite{huang2010multimodal} and Bayesian classification based on kurtosis and Skewness estimation \cite{hsia2008bayesian} for in-bed subject detection.

In recent years, there has been an increase in the number of studies on in-bed posture recognition using high density pressure sensors to estimate pose and limb locations. For instance, \cite{sun2016bed} used a $48\times128$ array of sensors for posture recognition based on weighted limbs, and \cite{boughorbel2012pressure} used $42\times192$ array of pressure sensors for classifying $4$ major sleeping positions of Supine, Prone, Left, and Right. Although these two approaches report a very high performance of over $95\%$ accuracy, they have used a very large number of sensors for their task. In contrast, \cite{viriyavit2017neural} used only two piezoelectric and pressure sensors for data collection. They reported near $90\%$ accuracy on $120$ hours of data recorded for only one subject. Similarly, \cite{rus2014recognition} reported $80.7\%$ accuracy by using $48$ conductive sensors. However, unlike the dataset used in our study \cite{pouyan2017pressure}, the number of posture classes in previous studies were limited to the few main sleeping positions, i.e. supine, left, and right side.

Multitask learning has been successfully used as a method of generalizing a classifier by learning multiple tasks at the same time. It has been used across many applications of machine learning, such as natural language processing \cite{collobert2008unified}, speech recognition \cite{deng2013recent} and computer vision \cite{zhang2014facial,li2016deepsaliency,zhang2016joint}. In these settings, by forcing the model to learn the shared structures of the appropriate tasks, the classifier benefits from each task and performs better than learning each task individually. In this paper, we also leverage the accuracy of our model by utilizing multitask deep learning model to predict both subject identities and their postures at the same time.

In this paper, we propose an end-to-end framework for in-bed posture and subject classification. Our method utilizes deep convolutional neural networks (CNN) to classify subjects and their sleeping posture from a single frame of data. We show that our method significantly outperforms other works including the state of the art. In our model, we introduce an additional hyper-parameter which acts as the coefficient of posture recognition and subject identification loss. We illustrate that by learning both tasks simultaneously and fine-tuning this parameter, our model is able to achieve significant performance gain. We confirm the effectiveness of our approach by incorporating a leave-one-subject-out (LOSO) evaluation scheme as well as $k$-fold evaluation, which was presented in previous studies on the same dataset \cite{pouyan2017pressure}. To further evaluate our system, we have introduced additional classification results by utilizing classical algorithms namely $k$-nearest neighbour ($k$NN) \cite{dudani1976distance}, support vector machines (SVM) \cite{suykens1999least}, and bagged trees \cite{breiman1996bagging}.

\begin{figure}[t]
\centering
\begin{center}
{\includegraphics[width=0.9\columnwidth]{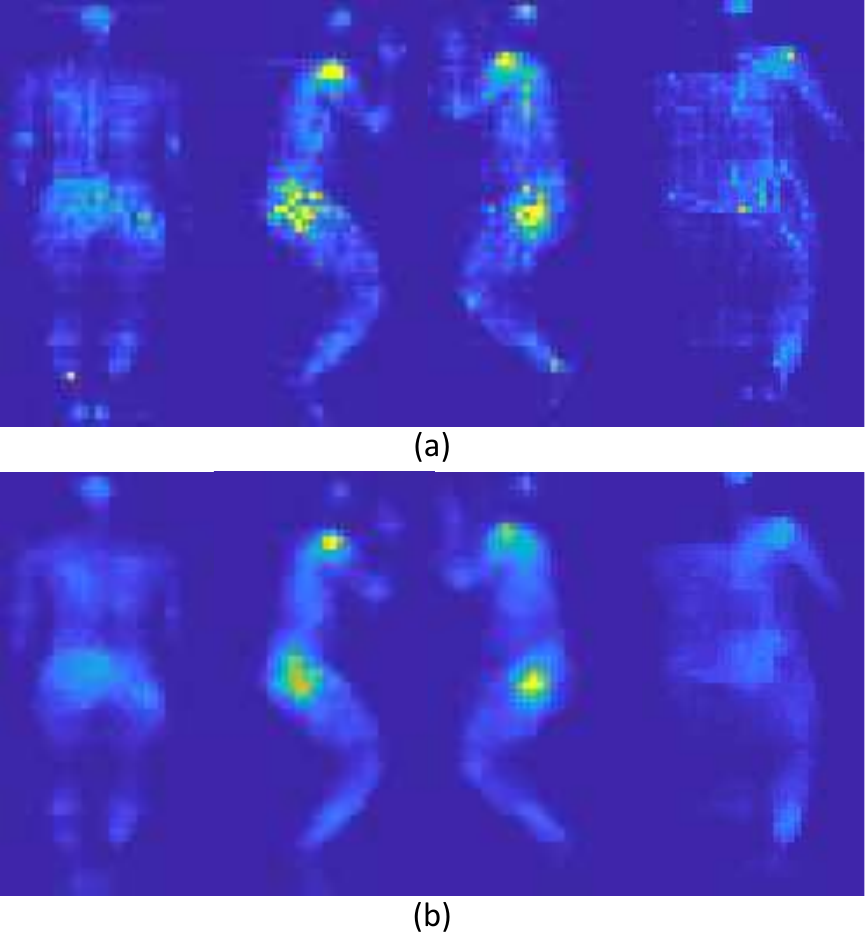}}
\end{center}
\vspace{-5mm}
\caption{Several examples of pressure maps from the dataset are illustrated. In (a), the samples are presented in raw format, while (b) illustrates the samples after pre-processing.}
\label{fig:fig1}
\end{figure}

\section{Material and Methods}
\subsection{Data and Pre-processing}
We used pressure map data provided in a public dataset, PmatData \cite{pouyan2017pressure, PhysioNet}, to train and test our pressure-based posture and user recognition system. The pressure data is collected using Vista Medical FSA SoftFlex $2048$. Each mattress contained $2048$, $1$ square inch, sensors placed on a $32\times64$ grid. The sensors report numbers in the range of $0-10000$. The data was collected at a sampling rate of $1Hz$ from $13$ participants in $8$ standard and $9$ additional uncommon postures. The age of participants was $19-34$ years, with a height and weight of $170-186$ $cm$ and $63-100$ $Kg$ respectively. A total of approximately $1800$ samples was recorded from each subject.

\begin{figure*}[t]
\centering
\begin{center}
{\includegraphics[width=\textwidth]{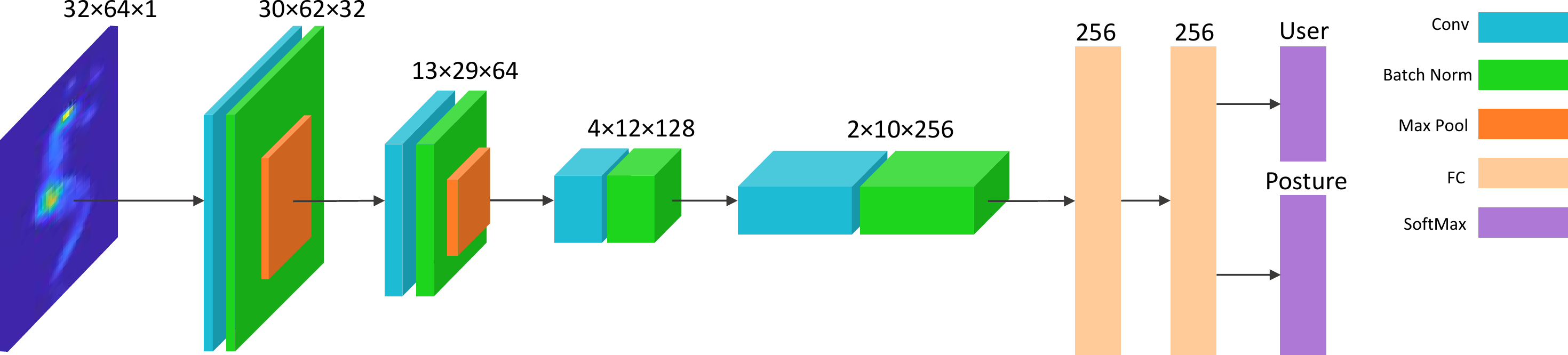}}
\end{center}
\vspace*{-4mm}
\caption{Our proposed framework is presented. A network consisting of four main blocks is designed to convert the pressure map data manifold into a feature space, which is then fed to two dense layers leading to two multinomial logistic regressors for simultaneous recognition of subjects and their sleeping postures.}
\label{fig:fig2}
\end{figure*}

In the pre-processing step we attempted to reduce the noise caused by occasional malfunctioning pressure sensors. Such artifacts often take place when individual sensors are subjected to large pressure values outside of the allowed voltage range. To cope with this problem, we applied a spatio-temporal median filter of size $3 \times 3 \times 3$ pixels in the image space and subsequently normalized the measurements. We also removed the first and last $3$ frames of each sequence, as in some cases they did not contain a clear image of the subject. Additionally, we removed $14$ data samples that contained no images. Some of the raw and pre-processed pressure maps are illustrated in Figure \ref{fig:fig1}.

\subsection{Classification}
We used a deep learning approach \cite{lecun2015deep} for subject and posture identification. As illustrated in Figure \ref{fig:fig2}, The classification is carried out by a feed-forward network which consists of two blocks of Conv-BatchNorm-MaxPool-LeakyRelu followed by another $2$ blocks of Conv-BatchNorm-LeakyRelu, capturing important properties of each single frame. The classification is then achieved by feeding the outcome to two dense layers followed by two SoftMax used in parallel to classify subjects and postures separately. For regularization, dropout is also used in the dense layers during training.

Let $I \in \mathbb{R}^{\textit{W} \times \textit{H}}$ be a dataset consisting of pressure maps from a set of users $\Gamma_I$ in a set of postures $\Delta_I$. The objective is to learn a function $\Phi(I; \theta) : \mathbb{R}^{\textit{W} \times \textit{H} \times \abs{\theta}} \rightarrow \mathbb{R}^\textit{2}$, where $\theta$ is the set of trainable parameters and $\Phi$ projects pressure maps $I$ onto a representation space where subjects and their sleeping postures are better separated. To address this problem, we consider training a CNN to discriminate between users and postures in $I$.

Formally, we consider a training set composed of tuples $(I, \gamma, \delta)$ where $I$ is the input pressure map and $\gamma$ and $\delta$ are the user and the corresponding posture respectively, that is, $\delta \in \Delta_I$ and $\gamma \in \Gamma_I$. Accordingly, we created a neural network with multiple layers, with the objective of discrimination between users and their sleeping postures. The last layer of the neural network consists of two multinomial logistic regressors, placed in parallel, with $M$ and $N$ units, where $M$ and $N$ are the number of users and sleeping postures in the dataset with $M = \abs{\Gamma_I}$ and $N = \abs{\Delta_I}$, and estimates $P(\gamma|I)$ and $P(\delta|I)$. To learn both functions, we consider using the multi-class cross-entropy loss of the users and their postures:
\begin{equation}
\begin{aligned}
\mathcal{L}_{user} & = -\sum_{j=1}^{M}{\gamma_{ij} \log P(\gamma_j | I_i)},\\ 
\end{aligned}
\end{equation}
\begin{equation}
\begin{aligned}
\mathcal{L}_{posture} & = -\sum_{j=1}^{N}{\delta_{ij} \log P(\delta_j | I_i)}.
\end{aligned}
\end{equation}

For training the model, we consider a loss function that combines the two losses and minimizes both at the same time. Our approach considers a weighted sum of both individual losses:

\begin{equation}
\begin{aligned}
\mathcal{L}& = \lambda\mathcal{L}_{user}+(1-\lambda)\mathcal{L}_{posture}, \label{equ:equ3}
\end{aligned}
\end{equation}
where $\lambda$ is a hyper-parameter that enforces the trade-off between the two mentioned objectives (separating users in $I$ and detecting postures). For this study, we trained our model with different values of $\lambda$ ranging from $0$ to $1$. Finally, for regularization purposes, we added the L2-loss of our convolution and FC layers' weights to the final loss.


In order to obtain a baseline for classification accuracy, we also incorporated several classical classifiers, namely SVM with a quadratic kernel, $k$NN with $k=10$, and bagged trees. Additionally, a multi-layer perceptron (MLP) neural network was utilized which consisted of $5$ layers of fully connected units, with a LeakyRelu activation functions, and $128, 256, 256, 128,$ and $64$ nodes in each layer. These algorithms are used frequently in IoT devices and smart home settings \cite{ozdemir2014detecting}. Therefore, this study provides a good example of how well they perform for practical smart beds. The parameters of these classifiers, such as $k$ in the $k$NN, the kernel in the SVM, and the number of hidden layers/nodes in the MLP, were selected and tuned empirically. The original features proposed for posture classification in \cite{pouyan2017pressure} were used in this phase.

\section{Experiments and Results}
Our proposed pipeline was implemented using TensorFlow on an NVIDIA Titan Xp GPU. The convolution kernels were $3 \times 3$ with a stride of $1$. Since the input pressure map dimension was small, bigger kernel sizes were not feasible with the required depth in the network. Up-sampling and padding did not contribute to the performance and therefore were not used. The convolutional layers were followed by MaxPooling with kernel size of $3 \times 3$, where applicable, and then followed by a batch normalization and LeakyRelu with negative activation coefficient of $0.2$. The pipeline then feeds into two dense layers with $256$ nodes each, with a dropout rate of $0.5$. Furthermore, each convolutional block was followed by an increasing dropout rate of $10\%$, $20\%$, $30\%$, $40\%$. The dropout layer acts as a regularizer to prevent the network from over-fitting by distributing the essential information between different nodes. We also employed an L2 regularization loss with a coefficient of $\sigma=0.002$. Finally, two SoftMax regressors were employed, one for classifying the subjects, and the other for recognizing their postures. We used an Adam optimizer at the training stage with a learning rate of $2\times10^{-5}$, which was decayed with a rate of $0.95$ every $10$ epochs. We trained the network for a total of $40$ epochs and a batch size of $64$. For our final model, we chose the best $\lambda$ for each validation scheme ($\lambda=0.5$ for $k$-Fold cross validation and $\lambda=0.2$ for LOSO).

\begin{figure}[t]
\centering
\begin{center}
{\includegraphics[width=0.9\columnwidth]{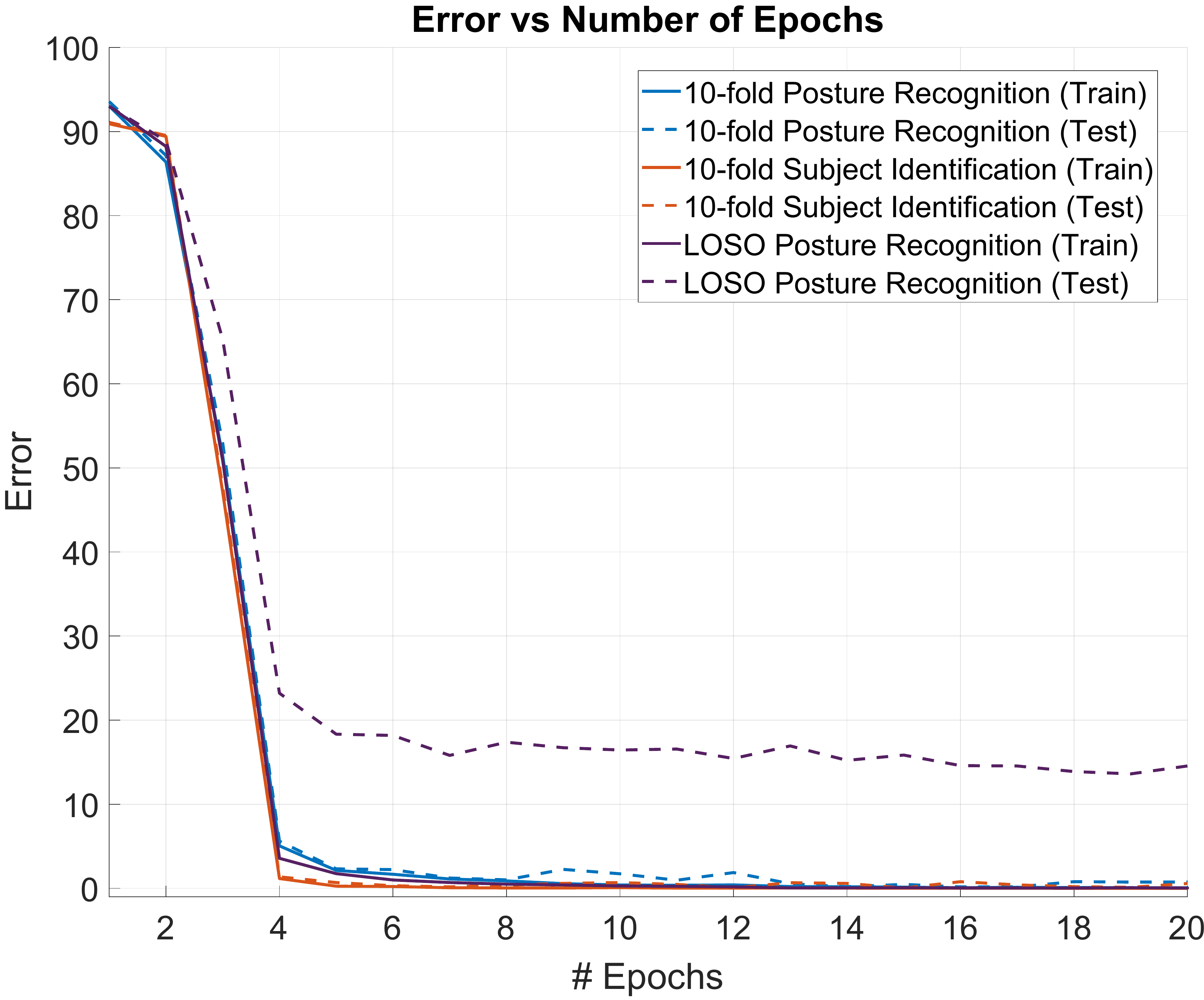}}
\end{center}
\vspace*{-3mm}
\caption{Task specific accuracy losses vs. the number of epochs for presented model are illustrated for our experiments.}
\label{fig:fig7}
\end{figure}

For validating our proposed method, we trained our model based on two validation schemes, namely, $k$-fold (with $k = 10$) and LOSO. In the $k$-fold cross-validation, we split the data into $10$ subsets, where $10\%$ was used for testing and $90\%$ for training. The process was repeated $10$ times. Finally, we averaged all the results, which resulted in a total posture and subject recognition accuracy of near $100\%$. In the LOSO scheme, one subject was kept aside at each iteration for testing and the rest of the subjects were used in training. Consequently, with LOSO, subject classification was not possible as evaluation needs to be done on the test subject.  However, the loss function provided in Equation \ref{equ:equ3} was used in both validation schemes regardless. Figure \ref{fig:fig7} presents the training and test curves of both validation schemes for posture recognition and subject identification, as well as $10$-fold cross-validation for subject identification. It can be seen that our model quickly learns and converges to a steady state, most likely due to the low variance (consecutive frames) in the data. Moreover, as expected, the test curve for the LOSO scheme exhibits higher error, which is due to the fact that the test frames are from a subject whose data has not been included in the training - a scenario which is not guaranteed for $10$-fold cross-validation.

\begin{table}
\centering
\caption{$10$-Fold Subject Identification Accuracy over $3$ Postures (in $\%$).} \label{table:confusion_fold}
\vspace*{-3mm}
\begin{center}
\footnotesize
\centering
\begin{tabularx}{\columnwidth}{cYYY}
\Xhline{2\arrayrulewidth} & Supine & Right & Left \\ \cline{1-4}
\footnotesize{Ref. \cite{pouyan2017pressure}} & 88.8 & 76.5 & 83.3 \\
\footnotesize{Proposed method \tiny{($\lambda$=$0.5$)}} & \textbf{99.9} & \textbf{100} & \textbf{100} \\ \Xhline{2\arrayrulewidth}
\end{tabularx}
\end{center}
\end{table}

We compared the performance of our model for subject identification across the $3$ general postures (supine, left side, and right side), with $10$-fold cross-validation, to \cite{pouyan2017pressure} which also used $k$-fold validation across the same $3$ main postures. The results are presented in Table \ref{table:confusion_fold}, where our model significantly outperforms \cite{pouyan2017pressure}. It is important to note that in \cite{pouyan2017pressure}, a separate model was trained for each of the three postures, and therefore, in Table \ref{table:confusion_fold}, we are showing the accuracy averaged over the three classifiers. 



Table \ref{table:confusion_loso} illustrates the performance of our model in classification of the $3$ main postures, evaluated using both $10$-fold and LOSO cross-validation. In order to further evaluate the performance of our model, we implemented $3$ baseline classification methods, namely SVM, $k$NN, and bagged trees. These classifiers were fine-tuned to provide the highest accuracy. It can be observed that, similar to our model, the baseline classifiers perform with high accuracy with $10$-fold validation. However, when faced with samples from subjects not involved in training (LOSO), the baseline methods experience a significant drop in accuracy, while our model performs robustly and almost performs the same as $10$-fold.

\begin{figure}[t]
\centering
\begin{center}
{\includegraphics[width=1\columnwidth]{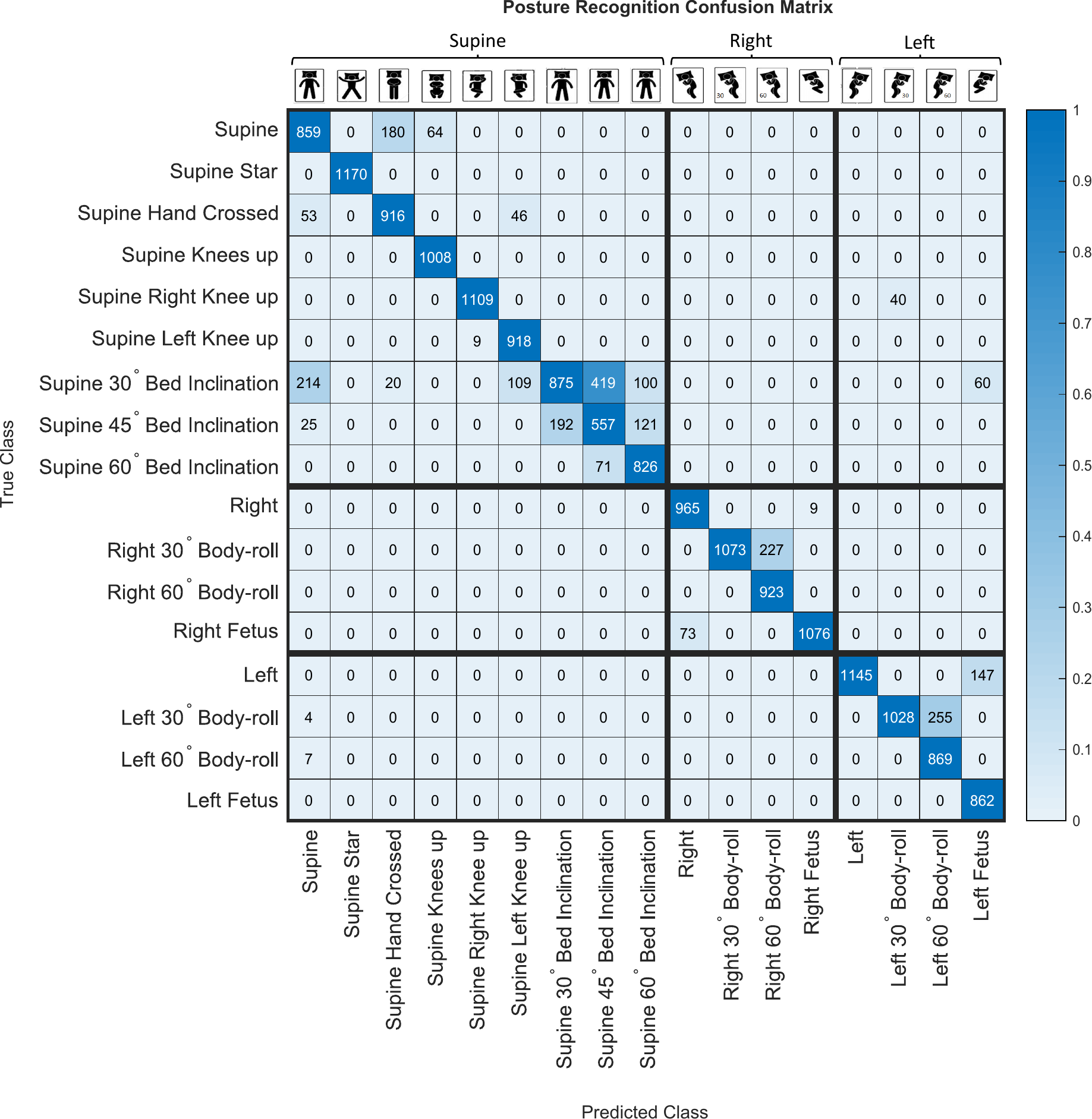}}
\end{center}
\vspace*{-3mm}
\caption{Posture recognition confusion matrix with leave-one-subject-out validation scheme. The posture categories are represented in 3 regions.}
\label{fig:fig3}
\end{figure}

As discussed above, our model performs very accurately in classification of the $3$ main sleeping postures. However, these postures themselves are composed of a number of sub-postures with very subtle differences. We trained our model on all these sub-classes (a total of $17$) and evaluated the performance using LOSO. Figure \ref{fig:fig3} presents the confusion matrix of this experiment. The average accuracy of our model for classification of these $17$ classes is  $87\%$, which, as expected, is lower than the $3$ main postures. However, a closer look at the confusion matrix reveals that most of the misclassified sub-posture fall within the correct \textit{main} posture (i.e. supine, left side, and right side). For instance, classes $1$ to $9$ are all minor variations of the supine posture, where in $7$, $8$, and $9$ only the bed is inclined by $30^{\circ}$, $45^{\circ}$, and $60^{\circ}$ respectively.


\begin{table}
\centering
\caption{Posture Recognition Precision (in \%).} \label{table:confusion_loso}
\vspace*{-3mm}
\begin{center}
\footnotesize
\centering
\begin{tabularx}{\columnwidth}{cYYY|YYY}
\Xhline{2\arrayrulewidth}
\footnotesize{Validation Scheme} &          \multicolumn{3}{c|}{10-Fold} &     \multicolumn{3}{c}{LOSO} \\ \cline{1-7} 
\footnotesize{Posture} & Supine & Right & Left & Supine & Right & Left \\ 
\cline{1-7} 
\footnotesize{Quadratic SVM} & 99.3 & 98.6 & 99.7 & 81.2 & 64.3 & 64.3 \\ \footnotesize{$k$NN ($k=10$)} & 99.9 & 99.6 & 99.9 & 75.6 & 46.1 & 54.3 \\ \footnotesize{Bagged Trees} & 99.8 & 99.9 & 99.9 & 90.6 & 54.0 & 65.0 \\
\footnotesize{Proposed method} & \textbf{100} & \textbf{100} & \textbf{99.9} & \textbf{99.0} & \textbf{100} & \textbf{99.7} \\
\Xhline{2\arrayrulewidth}
\end{tabularx}
\end{center}
\end{table}

During our experiments, we evaluated our model over different values of $\lambda$ to study it's effects on the outcome of our proposed pipeline. As we had already achieved near $100\%$ performance for both posture and subject identification in $k$-fold cross-validation, evaluation of $\lambda$ was only carried out in the LOSO scheme. Even though subject classification was not possible with the LOSO scheme, our experiments showed that by learning both posture and identity (only used in training), our model gains a performance improvement of $3.4\%$. We performed a t-test on accuracy for $\lambda=0$ and $\lambda\ne0$, to evaluate the significance of this difference, and achieved $p = 0.01$, indicating significant effect for $\lambda$.

\begin{table}
\centering
\caption{Series of data augmentation steps.} \label{table:augment}
\vspace*{-3mm}
\begin{center}
\footnotesize
\centering
\begin{tabularx}{\columnwidth}{cY}
\Xhline{2\arrayrulewidth}
Augmentation Probability & Process \\ \Xhline{2\arrayrulewidth}
$50\%$ & Rotation by $180^\circ$ \\ 
$20\%$ & Translation by up to $\pm10\%$ along $x$ \\ 
$20\%$ & Translation by up to $\pm10\%$ along $y$ \\ 
$20\%$ & Rotation by up to $\pm25^\circ$ \\  \Xhline{2\arrayrulewidth}
\end{tabularx}
\end{center}
\end{table}

Lastly, to further investigate the capacity of our model and its ability to generalize to more diverse datasets, we tested and trained our solution on augmented data. Data augmentation was done in a sequential process in such a way to resemble different body postures in the bed. Each step of data augmentation was assigned a probability according to the Table \ref{table:augment}, which shows the probability of each augmentation being applied. We trained using the augmented dataset with $\lambda=0.5$ with $50$ number of epochs. Our model achieved a subject identification accuracy of $89.7\%$ and posture recognition of $93.2\%$ for $17$ postures and $99.2\%$ for $3$ posture categories with the $k$-fold cross-validation scheme. Additionally, a posture recognition accuracy of $75.6\%$ over all $17$ classes and $98.2\%$ over the $3$ general sleeping pose categories with LOSO validation was achieved. These results show the robustness of our method even when used with more diverse sleeping postures.

\section{Conclusions}
In this work, a deep end-to-end CNN framework is proposed for ubiquitous classification of patient postures and identities with in-bed pressure-sensors. Our work is an efficient and highly accurate model that performs both tasks simultaneously, where conventional classification algorithms fail to achieve a good accuracy for unseen data. Our method can provide valuable information, which can be used in smart home and clinical settings. The proposed approach can predict subject's in-bed postures with an average accuracy of over $99\%$ based on LOSO evaluation scheme.  Results from data augmentation experiments suggest that our method can be generalized for an accurate estimation of even larger datasets. We also demonstrated the effects of the combined loss function, and by means of statistical tests, concluded that learning both tasks simultaneously can help improve the posture recognition performance. The future work can focus on pressure datasets with higher temporal resolution in order to measure physiological signals such as heart-rate and respiration-rate to further study the effects of sleeping postures on patient's state of health. 

\section*{Acknowledgment}
This work was funded by the Natural Sciences and Engineering Research Council of Canada (NSERC), Discovery Grant RGPIN-2018-04186. The Titan XP GPU used for this research was donated by the NVIDIA Corporation.

\bibliographystyle{IEEEtran}
\bibliography{IEEEabrv,main_bib}

\end{document}